\documentclass[letterpaper, 10 pt, conference]{ieeeconf}  

\IEEEoverridecommandlockouts                              

\usepackage{cite}
\usepackage{amsmath,amssymb,amsfonts}
\usepackage{algorithmic}
\usepackage{graphicx}
\usepackage{textcomp}
\usepackage{xcolor}
\usepackage{booktabs}
\usepackage{adjustbox}
\usepackage{color,soul}

\title{\LARGE \bf
Automated Measurement of Eczema
Severity \\
with Self-Supervised Learning
}

\author{Neelesh Kumar$^{1}$ and Oya Aran$^{1}$
\thanks{$^{1}$Neelesh Kumar and Oya Aran are with Corporate Functions-R\&D, The Procter and Gamble Company, Mason, OH, USA. 
        {\tt\small \{kumar.n.40, aran.o\}@pg.com}}%
}

\begin{document}

\maketitle
\thispagestyle{empty}
\pagestyle{empty}

\begin{abstract}
Automated diagnosis of eczema using images acquired from digital camera can enable individuals to self-monitor their recovery. The process entails first segmenting out the eczema region from the image and then measuring the severity of eczema in the segmented region. The state-of-the-art methods for automated eczema diagnosis rely on deep neural networks such as convolutional neural network (CNN) and have shown impressive performance in accurately measuring the severity of eczema. However, these methods require massive volume of annotated data to train which can be hard to obtain. In this paper, we propose a self-supervised learning framework for automated eczema diagnosis under limited training data regime. Our framework consists of two stages: i) Segmentation, where we use an in-context learning based algorithm called SegGPT for few-shot segmentation of eczema region from the image; ii) Feature extraction and classification, where we extract DINO features from the segmented regions and feed it to a multi-layered perceptron (MLP) for 4-class classification of eczema severity. When evaluated on a dataset of annotated "in-the-wild" eczema images, we show that our method outperforms (Weighted F1: 0.67 $\pm$ 0.01) the state-of-the-art deep learning methods such as finetuned Resnet-18 (Weighted F1: 0.44 $\pm$ 0.16)  and Vision Transformer (Weighted F1: 0.40 $\pm$ 0.22). Our results show that self-supervised learning can be a viable solution for automated skin diagnosis where labeled data is scarce. 
\end{abstract}

\section{Introduction}
Eczema is the most common skin disorder affecting over 10\% of the population in the United States alone. Previous studies have shown that allowing individuals to self-monitor their recovery can potentially lead to more effective recovery through positive psychological effects \cite{horne1989preliminary,ehlers1995treatment}. To that effect, a lot of work has been done on automated diagnosis of eczema from digital camera images, with emphasis on measuring the severity of eczema on a multi-point scale \cite{diepgen2009hand, agner2015classification, junayed2020eczemanet}.

The automated diagnosis is usually a two-step process: i) segmenting out the eczema region from the image \cite{nisar2023eczema,anand2023fusion}; ii) measuring the severity of eczema in the segmented region \cite{diepgen2009hand, agner2015classification, junayed2020eczemanet}. For both these steps, the state-of-the-art algorithms are based on supervised learning using deep neural networks (DNN) \cite{lecun2015deep}. In this paradigm, sufficient amount of labeled data is used to train the networks for segmentation and classification tasks. While these methods have shown impressive performance, acquiring a sufficiently large and diverse dataset is often difficult due to the variability of eczema across factors such as skin tone, severity levels, and anatomical location \cite{shi2019active}. Ensuring adequate representation for each of these variations can be time-consuming and costly, and therefore pose a hindrance in developing robust and generalizable DNN-based algorithms \cite{seth2017global,hurault2022detecting}.

In a previous work, we showed how the paradigm of in-context learning can be used for segmentation of eczema using just 2-3 training images \cite{kumar2024visual}. However, such a method for classification of eczema severity is lacking. The most popular method to train high-capacity networks using limited training data relies on transfer learning using finetuning \cite{tan2018survey,farahani2021brief}. A large network is first pretrained on a large source dataset, and then the classification heads of the pretrained network are retrained (or finetuned) on a target dataset. This simple technique of transfer learning has been quite successful in a lot of downstream image classification tasks. However, the method comes with a couple of issues that can limit the performance: i) there is still a requirement of enough labeled training data to finetune which can be hard to obtain and ii) the performance depends on how similar the target dataset is to the source dataset. If the difference is too high, then "transfer" of features from source dataset to target dataset can be challenging \cite{zheng2025learning}.

To address the limitations of traditional fine-tuning approaches, methods based on contrastive learning \cite{chen2020simple,caron2020unsupervised} and self-supervised learning (SSL) \cite{jaiswal2020survey,krishnan2022self,liu2021self} have been proposed. Contrastive learning techniques, such as SimCLR \cite{chen2020simple}, aim to learn robust representations by maximizing agreement between differently augmented views of the same image. In parallel, self-supervised learning methods like DINO (Distillation with No Labels) \cite{caron2021emerging} use a teacher-student framework where the student network learns to predict the output of the more slowly updated teacher network across different views of the same image, effectively distilling knowledge without labeled data. This approach allows DINO to capture rich and diverse features that generalize well across various tasks \cite{amir2021deep}. By leveraging these advanced SSL techniques, we can extract meaningful and task-relevant features from eczema images. This can allow us to perform more accurate severity classification and effectively overcome the challenges associated with limited labeled data.

\begin{figure*}[t!]
    \vspace{5.2pt}
    \centering
    \includegraphics[width=\textwidth]{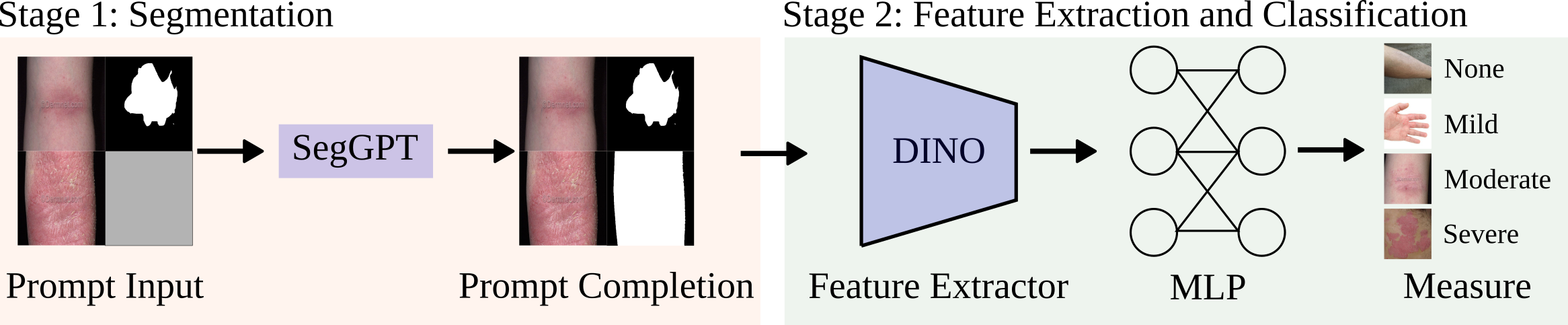}
    \caption{\textbf{Overview.} Our framework consists of two stages: In the first stage, we use a pretrained Seg-GPT model to perform segmentation of eczema region using just 2 labeled examples in the prompt. In the second stage, we use a pretrained DINO model to extract features from the segmented region. The features are then used to train an MLP to perform classification of eczema severity levels. }
    \label{fig:fig1}
\end{figure*}

In this work, we present a framework for automated eczema diagnosis relying on in-context learning and SSL, and thereby addressing the critical challenge of limited annotated data. Our framework consists of two stages: i) Segmentation, where we use an in-context learning based algorithm called SegGPT for few-shot segmentation of eczema region from the image and ii) Feature extraction and classification, where we extract DINO features from the segmented regions and feed it to a multi-layered perceptron (MLP) for 4-class classification of eczema severity. We evaluated the algorithm on a dataset of annotated "in-the-wild" eczema images and show that our method outperforms (Weighted F1: 0.67 $\pm$ 0.01) state-of-the-art finetuned Resnet-18 \cite{he2016deep} (Weighted F1: 0.44 $\pm$ 0.16)  and Vision Transformer (ViT) \cite{dosovitskiy2020image} (Weighted F1: 0.40 $\pm$ 0.22) methods. Our results highlight the potential of self-supervised learning in developing effective skin diagnosis applications, demonstrating that it can be a viable solution when labeled data is scarce.

\section{Methods}
We propose a two-stage framework for automated diagnosis of eczema from digital images. In the first stage, we employ the visual in-context learning framework proposed in \cite{kumar2024visual} for few-shot segmentation of eczema. In the second stage, we extract DINO features from the segmented eczema region and train an MLP using these features to predict the eczema severity level. 

\subsection{SegGPT for Eczema Segmentation}
We employ a pretrained Seg-GPT model \cite{wang2023seggpt} for few-shot segmentation of eczema region from the images. The segmentation task in SegGPT is formulated as an image in-painting task \cite{wang2023images}: given an input image, the prediction task is to inpaint the desired but missing output image. The model in SegGPT employs a vanilla vision transformer (ViT) \cite{dosovitskiy2020image} as the encoder consisting of stacked transformer blocks. A three layer head comprising of a linear layer, a 3 x 3 convolution layer, and another linear layer is used to map the features of each patch to the original resolution. A simple smooth $l_1$ regression loss is used to train the network. During inference, we use a $K$ nearest-neighbor algorithm to fetch $K$ images and their masks from the training set that are closest to the inference image. The task of the model is then to inpaint the mask for the inference image. 

\subsection{Self-Supervised Learning with DINO features}
In our framework, DINO \cite{caron2021emerging} plays a crucial role in extracting robust and transferable features from the segmented eczema regions. DINO is a self-supervised learning method that leverages knowledge distillation in the absence of labeled data, enabling the model to learn meaningful representations that generalize well across different tasks.

The DINO framework comprises two neural networks: a teacher network and a student network. Both networks are initialized with the same ViT architecture \cite{dosovitskiy2020image}, but they are updated differently during training: The teacher network parameters are updated using an exponential moving average (EMA) of the student network's parameters. This slow update mechanism provides stability and a consistent target for the student network. On the other hand, the student network is trained to predict the output of the teacher network for different augmented views of the same image. 

During training, both networks receive different views of the same inputs. The different views are generated through multiple augmentations, such as random cropping, flipping, color jittering, and Gaussian blurring. The output of the teacher network is mean-centered using the batch mean. The objective is for the student network's predictions to align closely with the teacher network's targets. The loss function used in DINO is a cross-entropy loss applied to the softmax outputs of the teacher and student networks. Specifically, the softmax output of the student network for each augmented view is compared against the softmax output of the teacher network for the same view. This encourages the student network to produce feature representations similar to those of the teacher network. 

\section{Experiments and Results}
The goals of our experiments were to investigate if leveraging DINO features can outperform traditional deep learning methods in the limited data regime. To that effect, we performed a comparative analysis of eczema severity classification using DINO features against ResNet-18 and ViT-B. 

\subsection{Dataset description}
Our dataset consists of 528 images collected from two primary sources: a) public Dermnet dataset of eczema images \cite{DermNet}, and b) a data collection study where images were taken directly by the participants using their smartphone cameras. The resulting dataset has multiple sources of variations: eczema images from different body parts, varying skin tones, and varying levels of eczema along with varying illumination, background, etc. which increases the complexity of the segmentation task. 

Each image in the dataset was labeled by human annotators into one of the four categories indicating overall severity of eczema: 0 indicating none; 1 indicating mild; 2 indicating moderate; and 3 indicating severe. The human annotators were provided written instructions from domain experts on how to assign these labels. Several examples were shown to them before they started the task. Each image was annotated by one human annotator. The resulting labels were then inspected by domain experts and were found to be satisfactory. 

All images were resized to 448 x 448 for the pretrained SegGPT model, and then to 224 x 224 for the DINO model. Further, the images were normalized using z-score normalization using the ImageNet dataset statistics. The dataset was partitioned into 5 random splits of 80\% train and 20\% validation images. 

\subsection{Baselines for comparison}
As baselines for comparison, we chose the widely adopted transfer learning technique of finetuning. We chose two classes of architecture: the convolutional-based ResNet18 \cite{he2016deep} and the attention-based ViT-B \cite{dosovitskiy2020image}. Both these models were pretrained on the ImageNet dataset. The classification heads of these pretrained networks were then finetuned on the target eczema dataset. 

The networks were trained on the training folds of each split using Adam optimizer with a small weight decay factor for additional regularization. Learning rate was set to $1e-4$. Batch size was set to $16$. The networks were trained for $50$ epochs to optimize the cross-entropy loss function. 

To evaluate, we use the weighted-averaged F1 score, which is computed by averaging the F1 scores across all classes, weighted by the number of samples in each class. The weighted-averaged F1 score balances precision and recall while accounting for class imbalance, providing a more comprehensive measure of classification performance compared to metrics like accuracy. The weighted-averaged F1 score was averaged across all the validation folds.

\subsection{Automated Eczema Diagnosis with the Proposed Framework}
In this section, we describe our proposed framework for automated eczema diagnosis, which consists of two main stages: segmentation of eczema regions and severity classification using self-supervised learning features.

\textbf{Stage 1: Eczema Region Segmentation}
In the first stage, we computed the segmentation masks indicating the eczema regions for all the images. We set $K=2$ and utilized the euclidean difference in the DINO feature space to retrieve the $K$ nearest neighbors for each inference image. This approach ensures that the segmentation task leverages contextual information from visually similar images, enhancing the accuracy of the segmentation masks.

\textbf{Stage 2: Feature Extraction and Classification}
Following the segmentation, we employed a pretrained DINO ViT-B model to extract feature representations from the segmented eczema regions. The DINO model, trained in a self-supervised manner, captures robust and invariant features without requiring labeled data. These extracted DINO features were then passed to a Multi-Layer Perceptron (MLP) for severity classification. The MLP architecture included one hidden layer of 128 dimensions, with dropout applied at a fraction of 0.3 for additional regularization to prevent overfitting.

\textbf{Training Setup}
We trained the MLP classifier using the Adam optimizer, optimizing the cross-entropy loss function for $50$ epochs. The learning rate was set to $1e-4$. The Adam optimizer was chosen for its adaptive learning rate properties, which help in achieving faster convergence and better performance.

We present the results of the comparison in Table \ref{tab:f1-performance}. Our method, with an average weighted F1 score of 0.67$\pm$0.01, outperforms both ResNet-18 (F1: 0.44$\pm$0.16) and ViT-B (F1: 0.40$\pm$0.22). The significantly high performance of our framework highlights its ability to effectively leverage self-supervised learning and in-context learning techniques for addressing the challenges associated with limited labeled data.

\begin{table}
\small
  \caption{Comparison for Eczema Severity Measurement}
  \label{tab:f1-performance}
  \centering
  \begin{tabular}{lc}
    \toprule
    Method & Weighted F1 Score \\
    \midrule
    \textbf{DINO} & \textbf{0.67$\pm$0.01} \\
    Resnet-18 & 0.44$\pm$0.16 \\
    ViT-B & 0.40$\pm$0.22 \\
    \bottomrule
  \end{tabular}
\end{table}

\begin{table}[b]
\small
  \caption{Effect of Segmentation on Performance}
  \label{tab:segmentation-effect}
  \centering
  \begin{tabular}{lc}
    \toprule
    Method & Weighted F1 Score \\
    \midrule
    DINO with segmentation & 0.67$\pm$0.01 \\
    DINO without segmentation & 0.65$\pm$0.03 \\
    \bottomrule
  \end{tabular}
\end{table}

\subsection{Ablation Study}

To emphasize the role of segmentation in improving classification performance, we conducted an experiment where the MLP was trained using DINO features extracted from the entire image instead of just the segmented region. The results, presented in Table \ref{tab:segmentation-effect}, indicate that while the improvement is marginal, incorporating segmentation leads to better classification performance. This suggests that focusing on the localized region of interest helps the model capture more relevant features, enhancing its ability to differentiate between severity levels.

\subsection{Limited Training Data Regime}
\begin{figure}[t!]
    \centering
    \vspace{5.2pt}
    \includegraphics[width=0.48\textwidth]{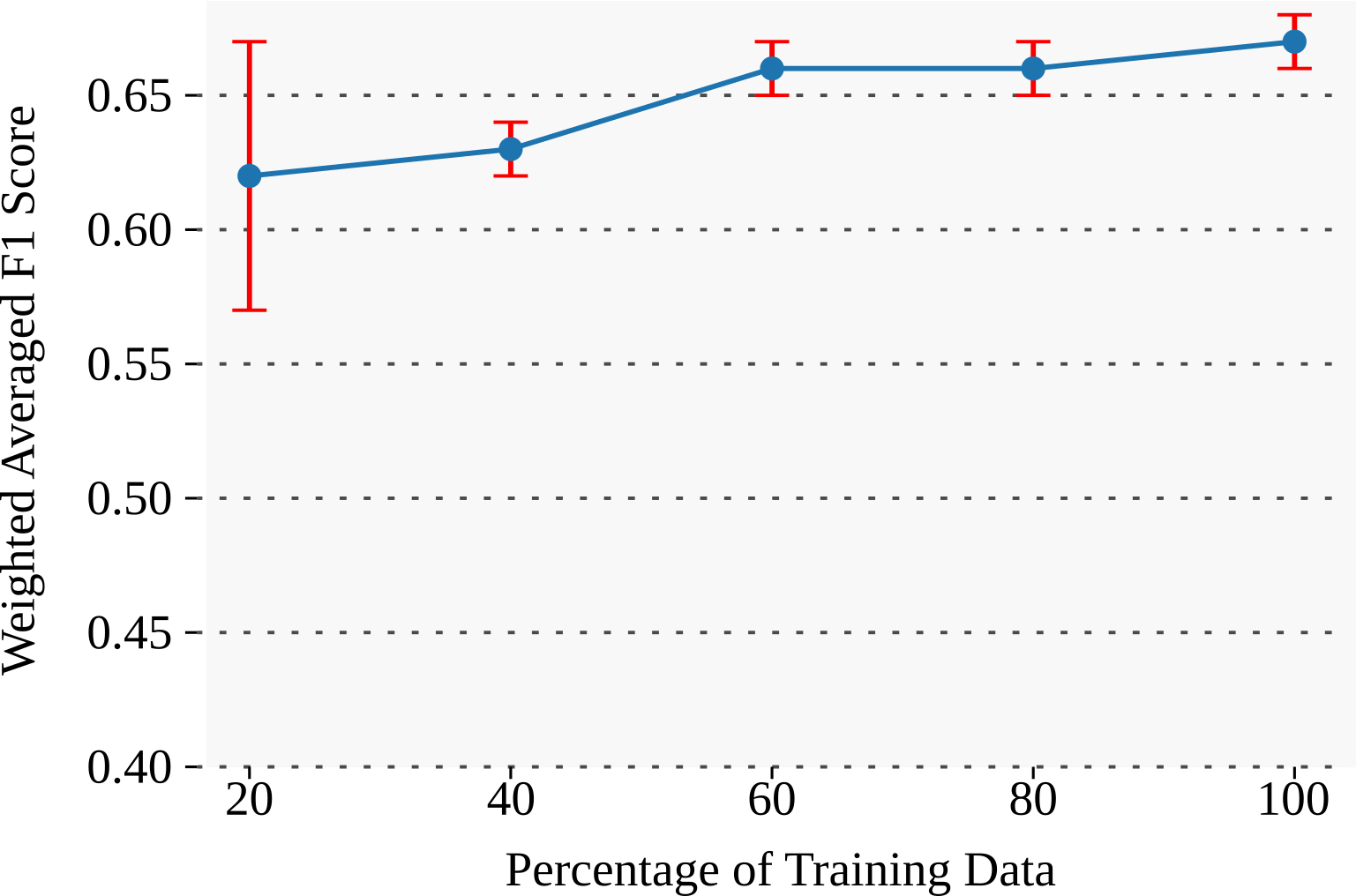}
    \caption{Training with limited data. Shown here are the means and standard deviations (red bars) for weighted averaged F1 score at different percentages of training data used. Even when trained with just 20\% of training data, the method exhibits only a slight decrease in performance. }
    \label{fig:fig2}
\end{figure}

To assess how the method performs with very limited training data, we conducted an experiment by training the models on progressively smaller subsets of the training data, ranging from 20\% to 100\% in 20\% increments. The validation set size was kept constant throughout the experiment. The results, shown in Figure \ref{fig:fig2}, demonstrate that even when trained with just 20\% of the available training data, the method exhibits only a slight decrease in performance. This highlights its robustness and effectiveness in low-data scenarios.

\section{Discussion}
In this work, we present a novel framework for the automated diagnosis of eczema from noisy, in-the-wild digital camera images. 
We demonstrated that by combining in-context learning for segmentation and self-supervised learning for classification, our method outperforms state-of-the-art finetuning techniques, especially in scenarios with limited labeled data. This reinforces the growing body of evidence suggesting that self-supervised ViT models such as DINO can indeed act as feature extractors encoding task-relevant local and semantic information \cite{amir2021deep}, and hence eliminating the need for extensive training or finetuning. 

Eczema is a highly variable condition, with its appearance influenced by factors such as skin tone, severity, and anatomical location \cite{seth2017global,hurault2022detecting}. Capturing this diversity in a training dataset is inherently challenging, particularly given the cost and time required for expert annotations \cite{shi2019active}. Existing methods often rely on supervised or transfer learning, but their performance degrades with insufficient or unbalanced datasets \cite{zheng2025learning}. Our approach bypasses many of these constraints by utilizing SSL to learn robust and generalized feature representations from unlabeled data, and an in-context learning-based segmentation model that operates effectively with as few as two labeled examples.

A key advantage of our framework is its applicability to real-world scenarios. For instance, individuals could self-monitor their condition using a digital device and only need to provide a small number of annotated examples to initialize the model. This could have profound implications for enabling individuals to track the progress of their eczema severity, fostering adherence to treatment regimens and offering a more personalized approach to managing their condition \cite{horne1989preliminary,ehlers1995treatment,wittkowski2007beneficial}.

Although SSL models such as DINO can extract task-relevant and meaningful features, there might still be room for improvement through further finetuning of DINO on domain-specific eczema datasets. This can enable it to handle subtle variations in severity and ensure robustness across diverse patient populations. Given the necessity of a large unlabeled dataset in finetuning DINO, future work could look at incorporating synthetic data to simulate underrepresented cases and further enhance classification performance \cite{ghorbani2020dermgan,sagers2023augmenting}.

In conclusion, our framework highlights the potential of integrating SSL and efficient segmentation techniques to address key challenges in automated eczema diagnosis. By reducing dependence on large-scale labeled data and providing scalable solutions for diverse real-world scenarios, this work points to a promising direction for AI-driven applications in dermatology. As SSL and related technologies continue to evolve, their role in personalized, data-efficient healthcare solutions is likely to expand, bringing new opportunities for clinical efficacy.

\bibliographystyle{splncs04}

\end{document}